\documentclass{article}

     \usepackage[preprint,nonatbib]{neurips_2021}

\usepackage[utf8]{inputenc} 
\usepackage[T1]{fontenc}    
\usepackage{hyperref}       
\usepackage{url}            
\usepackage{booktabs}       
\usepackage{amsfonts}       
\usepackage{nicefrac}       
\usepackage{microtype}      
\usepackage{xcolor}         

\usepackage{mathtools}

\newcommand{\irow}[1]{
  \begin{smallmatrix}(#1)\end{smallmatrix}%
}

\usepackage{graphicx}

\title{Double Descent Optimization Pattern and Aliasing: Caveats of Noisy Labels}

\author{%
  Florian Dubost\\
  Stanford University\\
  \texttt{floriandubost1@gmail.com} \\
  \And
  Erin Hong \\
  Stanford University\\
  \And
  Max Pike \\
  Stanford University\\
  \And
  Siddharth Sharma \\
  Stanford University\\
  \And
  Siyi Tang \\
  Stanford University\\
  \And
  Nandita Bhaskhar \\
  Stanford University\\
  \And
  Christopher Lee-Messer \\
  Stanford University\\
  \And
  Daniel Rubin \\
  Stanford University\\

}

\begin{document}

\maketitle

\begin{abstract}
Optimization plays a key role in the training of deep neural networks. Deciding when to stop training can have a substantial impact on the performance of the network during inference. Under certain conditions, the generalization error can display a double descent pattern during training: the learning curve is non-monotonic and seemingly diverges before converging again after additional epochs. This optimization pattern can lead to early stopping procedures to stop training before the second convergence and consequently select a suboptimal set of parameters for the network, with worse performance during inference. In this work, in addition to confirming that double descent occurs with small datasets and noisy labels as evidenced by others, we show that noisy labels must be present both in the training and generalization sets to observe a double descent pattern. We also show that the learning rate has an influence on double descent, and study how different optimizers and optimizer parameters influence the apparition of double descent. Finally, we show that increasing the learning rate can create an aliasing effect that masks the double descent pattern without suppressing it. We study this phenomenon through extensive experiments on variants of CIFAR-10 and show that they translate to a real world application: the forecast of seizure events in epileptic patients from continuous electroencephalographic recordings.
\end{abstract}

\section{Introduction}

The evolution of the generalization error during training is often closely analyzed by machine learning practitioners to make substantial decisions about hyperparameter tuning, model architecture, or collection of additional data. The estimation of the generalization error can consequently be considered to be at the core of machine learning.

Knowing when to stop training neural networks is crucial to reaching optimal generalization performance. Estimating the exact optimal stopping time is still subject to debate, with novel early-stopping strategies continuously being proposed in the literature \cite{dodge2020,song2019}. Most early stopping strategies would assume a steady increase in the generalization error across epochs as a reason to stop training. 

On new, unresolved tasks, networks may not have access to enough input information to predict the labels correctly. In this scenario, most machine learning practitioners would also stop training after observing a divergence of the generalization error, and potentially modify their research direction. For example, this can occur in medical datasets when the acquired patient data could be insufficient to identify the target outcome.

In most cases, most early stopping algorithms and machine learning practitioners would assume a steady increase in the generalization error across epochs as a reason to stop training. However, research has shown that small and noisy datasets may trigger an epoch-wise double descent optimization pattern \cite{nakkiran2019}. This means that after getting worse through the epochs, the generalization error reduces again, potentially leading to an overall smaller generalization error. Stopping the training before this second descent would be suboptimal.

Double descent has been qualitatively defined by others \cite{nakkiran2019}. In this article, we mathematically define epoch-wise double descent and confirm earlier findings demonstrating that double descent appears when the training set is both small and has noisy labels. We also empirically show that:
\begin{itemize}
\item Both training and generalization sets must have noisy labels for a double descent pattern to appear.
\item If only the labels of the training set are noisy, double descent does not occur. Instead, a plateau pattern may appear.
\item Even when the training set is small and labels are noisy in both training and generalization sets, there exists a learning rate for which the double descent pattern does not appear.
\item The double descent pattern appears when the learning rate is too small, with the exact value of the learning depending on the dataset and task.
\item Increasing the learning rate may create an aliasing effect that hides the double descent pattern without suppressing it.
\end{itemize}

\section{Related Work}

A recent article analyzing double descent patterns was published by a group from OpenAI \cite{nakkiran2019}. The authors investigate three types of double descent patterns: epoch-wise double descent patterns, which we also study; model-wise double descent patterns, where the generalization error is computed as a function of the model size; and sample-wise non-monotonicity, where the generalization error is computed as a function of the training set size. They establish that double descent patterns occur for large models trained on small datasets with noisy labels. They also establish a complexity measure called “effective model complexity” that indicates whether double descent will happen. However, this measure does not completely solve the practical problem of double descent, as it is sensitive to some of its hyperparameters. The major finding of the article is that in the double descent setting, more--but still not enough--data can make the model worse. The authors also hypothesize that the first local minimum of the generalization error in the double descent patterns corresponds to an overparametrized model memorizing the training set, before learning the true relationship between inputs and outputs.

Belkin et al. \cite{belkin2019} were the first to use the \textit{double descent} terminology. They experiment with Random Fourier Features, a type of neural networks, and with boosted stumps and random forests using the MNIST dataset \cite{lecun-mnisthandwrittendigit-2010}. They give insights that double descent occurs in overparameterized settings, which is in line with the findings of Nakkiran et al. \cite{nakkiran2019}.

Earlier, others \cite{opper1995,opper2001,advani2020,spigler2018,geiger2019} had also recognized non-monotonic patterns in the optimization of machine learning models. For example, Geiger et al. \cite{geiger2019} analyze how overparametrized settings of neural networks influence the optimization landscape and compare it to the energy landscape of repulsive ellipses.

Following the publication of Belkin et al. \cite{belkin2019}, other authors \cite{belkin2020,hastie2019,bartlett2020,muthukumar2020,bibas2019,mitra2019,mei2019} reused the double descent terminology. Mei et al. \cite{mei2019} underline the importance of regularization in overparametrized regimes and compare the influence of data augmentation, $l_{1}$ regularization and $l_{2}$ regularization on the optimization landscape.

Although the exact causality of double descent is still unknown, most authors seem to agree with the memorization hypothesis, which states that the first local minimum of the generalization error in the double descent patterns corresponds to an overparametrized model memorizing the training set.

\section{Methods}

A few studies have identified factors leading to double descent patterns. However, to the best of our knowledge, double descent patterns have only been qualitatively defined (as a steady increase of the generalization error preceding a steady decrease of the generalization error). We propose a quantitative definition of the double descent pattern based on the derivative of the generalization error, and propose to estimate those derivatives using polynomial fitting and the Vandermonde matrix.
Subsequently, we use this characterization of the double descent pattern to identify under which circumstances the double descent pattern occurs, and how it can be suppressed.

\subsection{Definition of a Double Descent Pattern}

Let us consider the generalization error $E : \mathbb{R} \mapsto \mathbb{R}$ as a function of the training time $t$. Given two times $t_{i}$ and $t_{j}$, and the generalization error $E$ being differentiable on the interval $[t_{i},t_{j}]$, we define a time-wise double descent pattern to be visible in the segment $[t_{i},t_{j}]$ if and only if the equation

\begin{equation}
\frac{\partial E(t)}{\partial t} = 0
\end{equation}

has at least one solution $t_{s}$ which satisfy the following property: there exists two epochs $t_{a}$ and $t_{b}$ such that $\frac{\partial E(t)}{\partial t} > 0$ for $t \in [t_{a},t_{s}]$, and $\frac{\partial E(t)}{\partial t} < 0$ for $t \in [t_{s},t_{b}]$. 

\subsection{Polynomial Approximation}

In practice, the generalization error $E$ is unknown and is estimated using an independent set of samples and labels. During training, the estimated generalization error $\hat{E}$ is only evaluated discretely in a finite number of $N$ epochs $e_{n}$ (which can be mapped to the training time using a linear function $f$). Consequently, the estimated generalization error $\hat{E}$ is not continuous – hence not differentiable. 

In order to compute derivatives, we propose to compute a differentiable estimate of the generalization error based on the $N$ epoch-wise estimates $\hat{E}(e_{n})$, with $f(\hat{E}(e_{n})) \in [t_{i},t_{j}]$. Specifically, we propose to fit a $k^{th}$ degree polynomial $P: t \mapsto a_{0}+a_{1}t+...+a_{k}t^{k}$ using least squares fitting. Consequently, we need to minimize the quantity $\sum_{n=0}^{N} |P(e_{n}) - \hat{E}(e_{n})|^2$. To solve this minimization problem, we can compute the polynomial coefficients as $\bold{a} = (\bold{V}^{T}\bold{V})^{-1}\bold{V}^{T}\bold{E}$, where $\bold{a}=\irow{a_{0}&  a_{1}& … &a_{k}}^{T}$, $\bold{E}=\irow{\hat{E}(e_{0}) & \hat{E}(e_{1})& …& \hat{E}(e_{n})}^{T}$, and $\bold{V}$ is a $ n \times (k+1)$ Vandermonde matrix \cite{turner1966,klinger1967} of the $N$ epochs $e_{n}$ for which $\det(V^{T}V) \neq 0$. 

Based on empirical evaluation, we recommend choosing the polynomial order $k$ in $[3,6]$: high enough to allow the fitting of double descent patterns, low enough to avoid overfitting to the noise in the estimation of the generalization error.

The times $t_{s}$ and $t_{b}$ should also be selected far enough from each other to allow a sufficient sampling of the generalization error, i.e such that $N$ is large enough. The choice of the distance between $t_{s}$ and $t_{b}$ could be compared to stopping criteria of early stopping procedures \cite{dodge2020,song2019}. Besides, if the interval $[t_{i},t_{j}]$ is undersampled, aliasing effects can hide the double descent pattern, and fail the proposed double descent detection methodology.

\subsection{Savitzky-Golay Smoothing}
If the epoch-wise estimated generalization $\hat{E}$ is too noisy, we propose to first use a Savitzky-Golay smoothing filter \cite{press1990} before applying polynomial fitting. The filter is a weighted moving average of polynomials, which are least-squares fitted within a given window size. 

\section{Experiments and Results}

We present a series of experiments designed to identify factors influencing the apparition of double descent patterns. Most of the experiments are trained for longer than shown in the figures. In order to visually compare the plots, we restrict the display boundaries. The full training curves are given in supplementary materials. 

\subsection{Tasks}
The experiments are performed for two tasks: image classification using a subset of the CIFAR-10 \cite{krizhevsky2009} dataset, and seizure forecasting from electroencephalograms using a private in-house dataset. We reframe both tasks as balanced binary classification tasks, use binary cross-entropy \cite{mackay2003} as our loss function and generalization error, and report results for two optimizers: Adam \cite{kingma2014} and Adadelta \cite{zeiler2012}. Currently, Adam is the most widely used optimizer, while Adadelta was designed to be almost insensitive to the learning rate. We describe below the datasets and corresponding networks.

\subsubsection{Image Classification with CIFAR-10}
\label{sec:cifar10}

CIFAR-10 is a dataset of 60,000 32x32 colour images in 10 classes, with 6,000 images per class \cite{krizhevsky2009}. To easily observe double descent patterns, we restrict the problem to a binary classification between the airplane and ship classes using only 300 images for training and 300 other images to evaluate the generalization error. Images were sampled at random such that there were 150 positive images (airplane) and 150 negative images (ship) in each set. We also created \textit{noisy} variants of the training and generalization (validation) sets, where 30 random positive images were swapped with negative images, without changing their labels. We create a last variant with 3,000 images--instead of 300--in the training set to analyze the effect of training set size on the apparition of double descent patterns. Unless mentioned otherwise, results are reported for the initial variant: 300 images and noisy labels.

For preprocessing, image intensity values are rescaled in $[0, 1]$ using the image-wise minimum and maximum to facilitate the training. We use a shallow ResNet-like \cite{he2016} network with two series of two $3 \times 3$ convolutional layers, with 32 kernels for the first series and 64 for the second, each followed by ReLU activations and separated by a $2 \times 2$ maxpooling layer. After the second series of convolutional layers, a global average pooling layer and a fully connected layer map to a single neuron output, on which sigmoid is applied. The network has 67,396 parameters.

\subsubsection{Seizure Forecasting from Electroencephalograms}

Electroencephalograms (EEGs) are continuously recorded from hospital patients with epilepsy. Times of seizure onsets and other unrelated events are indicated by clinicians in free text. Patient data collection for this study was approved by the Research Board of our institution under protocol IRB-37949. A waiver of consent was granted based upon the findings of minimal risks to patient welfare or rights and the impracticality of obtaining consent retrospectively for thousands of patients in studies occurring over many years. Personal health information is included neither in this article nor in the supplementary materials.

We process the text labels to create seizure time labels, which are consequently noisy by design. In the literature \cite{usman2020,meisel2019,truong2018}, seizure onset is predicted from EEG by discriminating between preictal signal--the signal that precedes seizure onset--and interictal signal--the signal that is far from seizure onset. Replicating values in the literature \cite{usman2020}, we assume the duration of the preictal time to be 30 minutes, and sample interictal signal after 35 minutes of seizure onset (5 minutes of seizure times plus 30 minutes of postictal time). In practice, the length of preictal, postictal and ictal times depend on the patient and seizure type. Consequently, our approximations are introducing additional noise in the labels. 

We select the nine patients who have the largest number of precital and interictal segments. In total, the gathered dataset comprises 366,119 seconds of interictal time and 123,185 seconds of preictal time, from 154 preictal segments and 182 interictal segments. This breaks down on average to 40,679 seconds of interictal time per patient, 14,353 seconds of preictal time per patient, 17 preictal segments per patient, 20 interictal segments per patient. We use only signals from the 19 electrodes in the standard 10–20 International EEG configuration, sampled at 200Hz. We extract 5 second long clips and compute the following 20 features for each electrode of each clip: mean, variance, standard deviation, peak-to-peak amplitude, skewness, kurtosis, Hurst exponent, approximate entropy, sample entropy, decorrelation time, Hjorth mobility, Hjorth complexity, Higuchi fractal dimension, Katz fractal dimension, number of zero-crossing, line length, spectral entropy, SVD entropy, SVD Fisher information, and spectral edge frequency. Those features are computed with the MNE library version 0.1, which also gives details about their computation \cite{schiratti2018}. Consequently, the dimension of each clip was $19 \times 20$, where 19 is the number of electrodes and 20 the number of features. Each clip is labeled as either interictal or preictal, and a neural network is trained to discriminate them. The rationale behind this approach is that having enough preictal segments detected during inference indicates that a seizure is imminent.

Before training, the feature values $x$ are logarithmically rescaled as $\log(1+x)$. We use multilayer perceptron of two layers with a first fully connected layer of 512 neurons followed by a ReLU activation, a second fully connected layer of 1024 neurons also followed by a ReLU activation and a final layer which maps to a single output neuron followed by a sigmoid activation. The network has 721,409 parameters.

The generalization error on unseen subjects did not converge when we attempted to optimize the networks on the full dataset. Instead, we decided to optimize the networks patient-wise. The split between training and generalization set was consequently done at random, patient-wise, on the segment level. This observation matches the clinical reasoning that the biological processes which trigger seizures may be patient-specific.

\subsection{Technical specifications}
Experiments are performed using two NVIDIA GTX 1070 GPUs and one TITAN RTX GPU on local machines. Training time varies from 30 minutes to two days depending on the dataset, optimizer and learning rate. The experiments on CIFAR-10 are substantially faster than those on seizure forecasting. Seizure forecasting experiments are realized with TensorFlow 2.4.0 \cite{tensorflow2015} and Keras 2.4.3 \cite{chollet2015}. Experiments on CIFAR-10 are realized with TensorFlow 1.1.0  \cite{tensorflow2015} and Keras 2.2.0 \cite{chollet2015}.

\subsection{Influence of the training set size on double descent patterns}

\begin{figure}[h!]
  \centering
  \includegraphics[width=14cm]{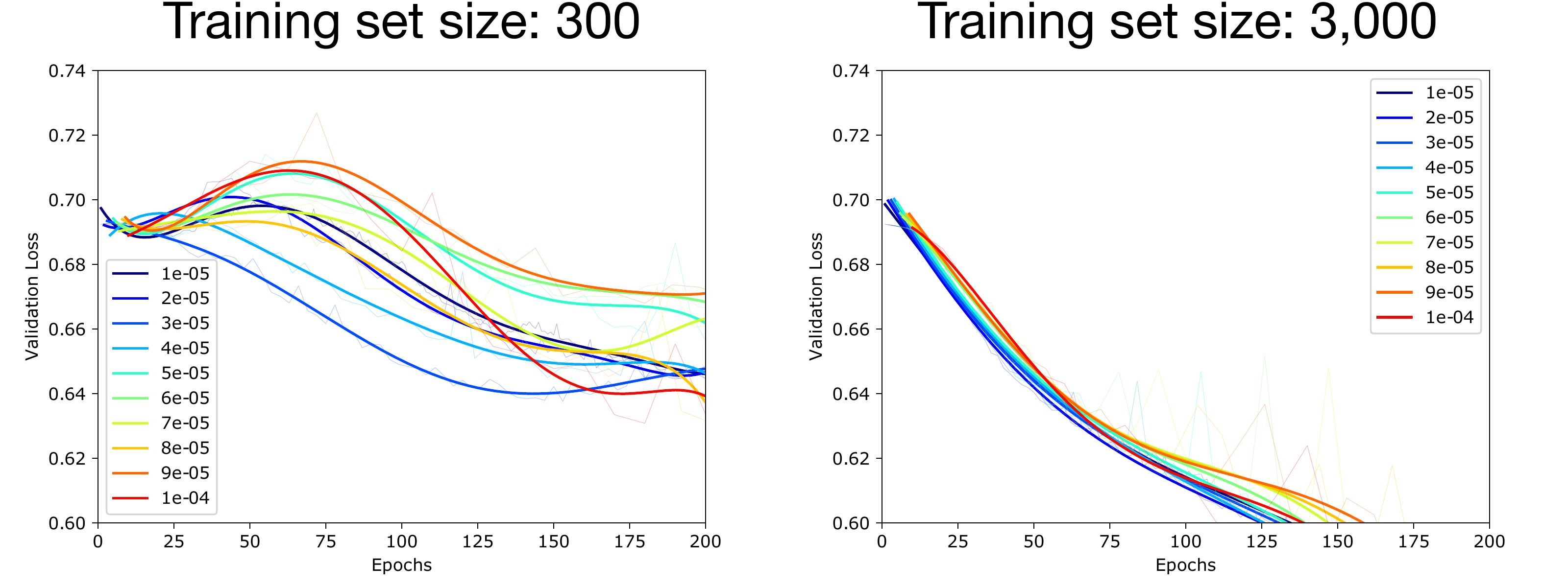}
  \caption{Influence of the dataset size on the apparition of a double descent pattern. On the left, the training set comprises 300 images, while on the right it comprises 3,000 images. For both plots, the generalization set comprises 300 images. Results are shown for values of the learning rate of the Adam optimizer (see legend). Polynomials were fitted to each estimated generalization error curve. The generalization error is indicated on the y-axis as \textit{validation loss}. Plots with more epochs are given in supplementary materials.}
  \label{fig:trainsetsize}
\end{figure}

One of the intuitions for the apparition of a double descent pattern is that the dataset is small w.r.t. the model size, which allows the model to rapidly memorize the training data \cite{nakkiran2019}. In our experiment on CIFAR-10, a double descent appears with a model of 67,396 parameters and only 300 images in our training set with noisy labels. Increasing the training set to 3,000 images--while keeping the same percentage of incorrectly labeled images (300 images)--makes the double descent pattern disappear (Figure \ref{fig:trainsetsize}). These results seem to confirm the memorization hypothesis \cite{nakkiran2019}. 

\begin{figure}[b!]
  \centering
  \includegraphics[width=11cm]{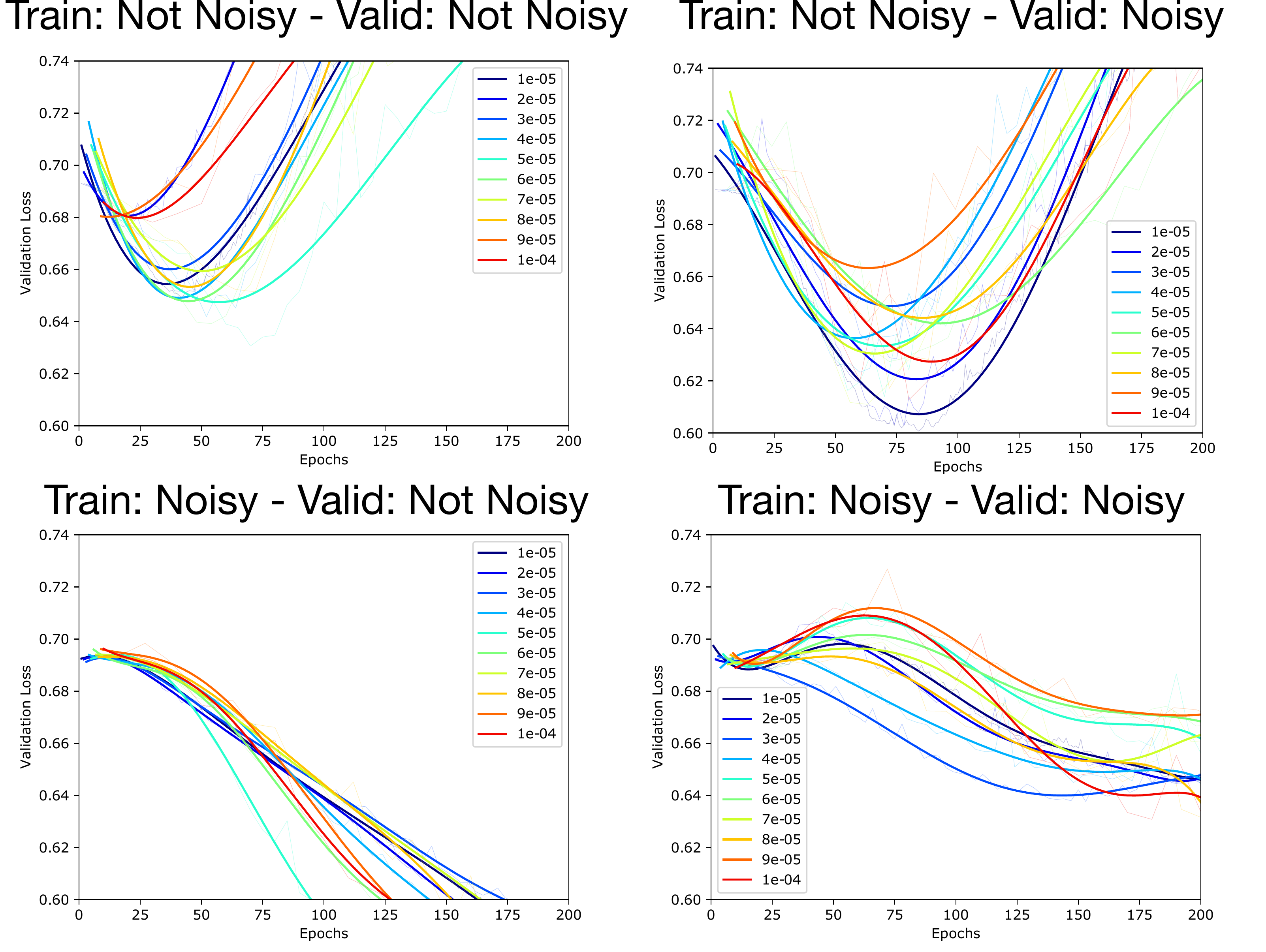}
  \caption{Influence of label noise on the apparition of a double descent pattern. On the top left subplot, the training and generalization (valid) set Results are shown for values of the learning rate of the Adam optimizer (see legend). Polynomials were fitted to each estimated generalization error curve. The generalization error is indicated on the y-axis as \textit{validation loss}. Plots with more epochs are given in supplementary materials.}
  \label{fig:trainVSvalid}
\end{figure}

\subsection{Influence of the label noise on double descent patterns}

Earlier studies found that double descent patterns appear when the dataset has noisy labels \cite{nakkiran2019}. We perform a series of experiments on CIFAR-10 using the noisy and clean versions of our datasets (Section \ref{sec:cifar10}). We investigate four scenarios: both the training and generalization sets are clean, both the training and generalization sets are noisy, the training set is noisy and the generalization set is clean, and the opposite: the training set is clean and the generalization set is noisy (Figure \ref{fig:trainVSvalid}). Our results confirm earlier findings that double descent appears in noisy datasets. We are also able to specify further that both the training and generalization sets must have noisy labels for a double descent to appear. As a side note, if the training set only has noisy labels, we may still observe a plateauing of the generalization error at the beginning of the training, which may also mislead early stopping algorithms into declaring the training process as finished. Our takeaway is that if all labels are noisy, one must expect double descent patterns to appear and adjust their early stopping criteria to be less conservative.

\begin{figure}[b!]
  \centering
  \includegraphics[width=12cm]{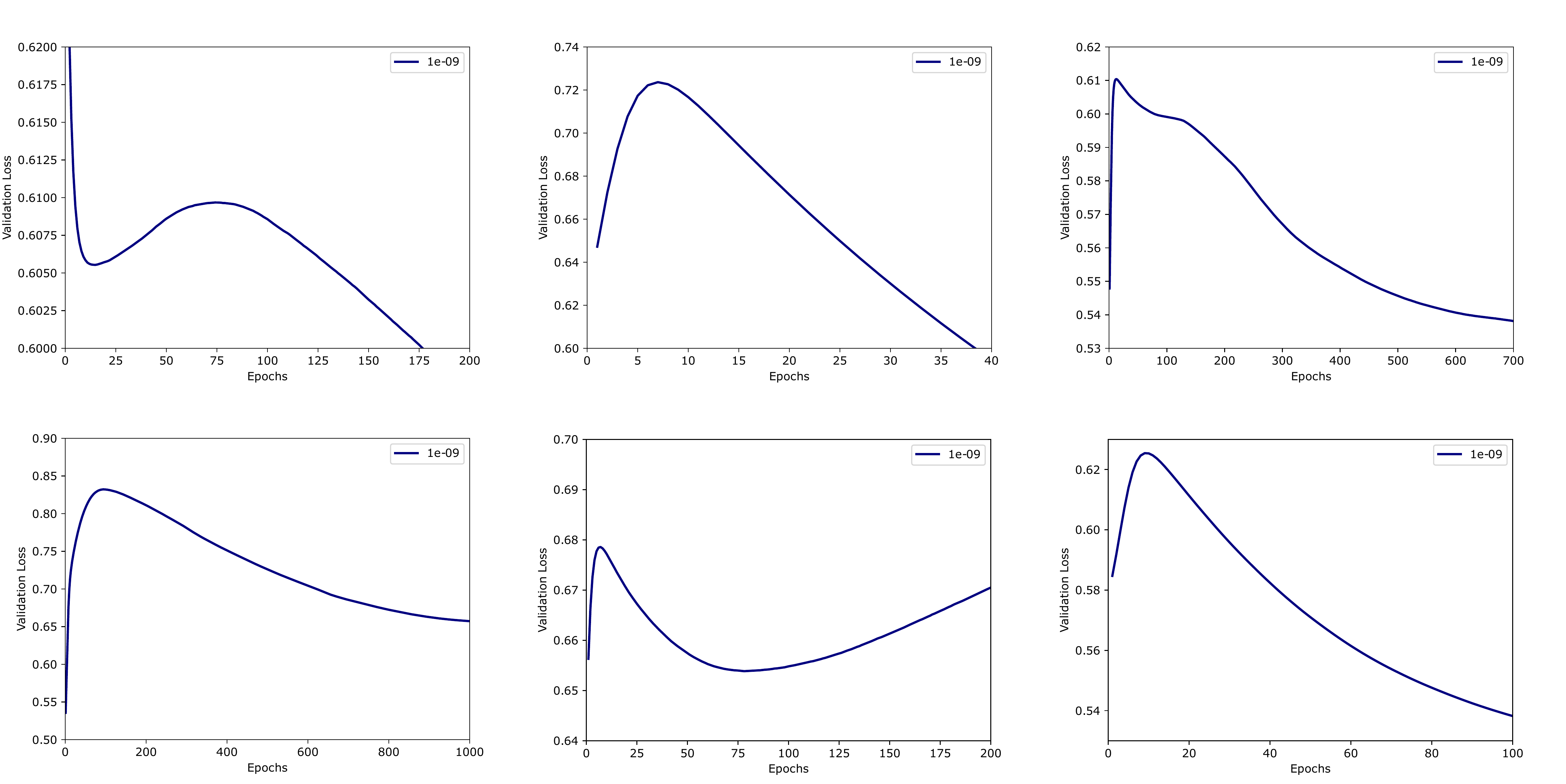}
  \caption{Double descent pattern visible for multiple patients in seizure forecasting. There is no polynomial fitting in this figure. The generalization error is indicated on the y-axis as \textit{validation loss}. The boundaries of the x-axis differ per subplot. All experiments were performed with Adam with a learning rate of 1e-9.}
  \label{fig:multisub_sz}
\end{figure}

\subsection{Real-life application: seizure forecasting}

Double descent patterns have mostly been analyzed theoretically on toy datasets. In addition to the experiments on the toy CIFAR-10 dataset, we also observe double descent in a real-life application: seizure forecasting from electroencephalograms. Out of 9 subjects, six display a double descent pattern during training (Figure \ref{fig:multisub_sz}). For one of the remaining subjects, the generalization error smoothly converges, and for the two other remaining subjects, the generalization error diverges.

\begin{figure}[b!]
  \centering
  \includegraphics[width=14cm]{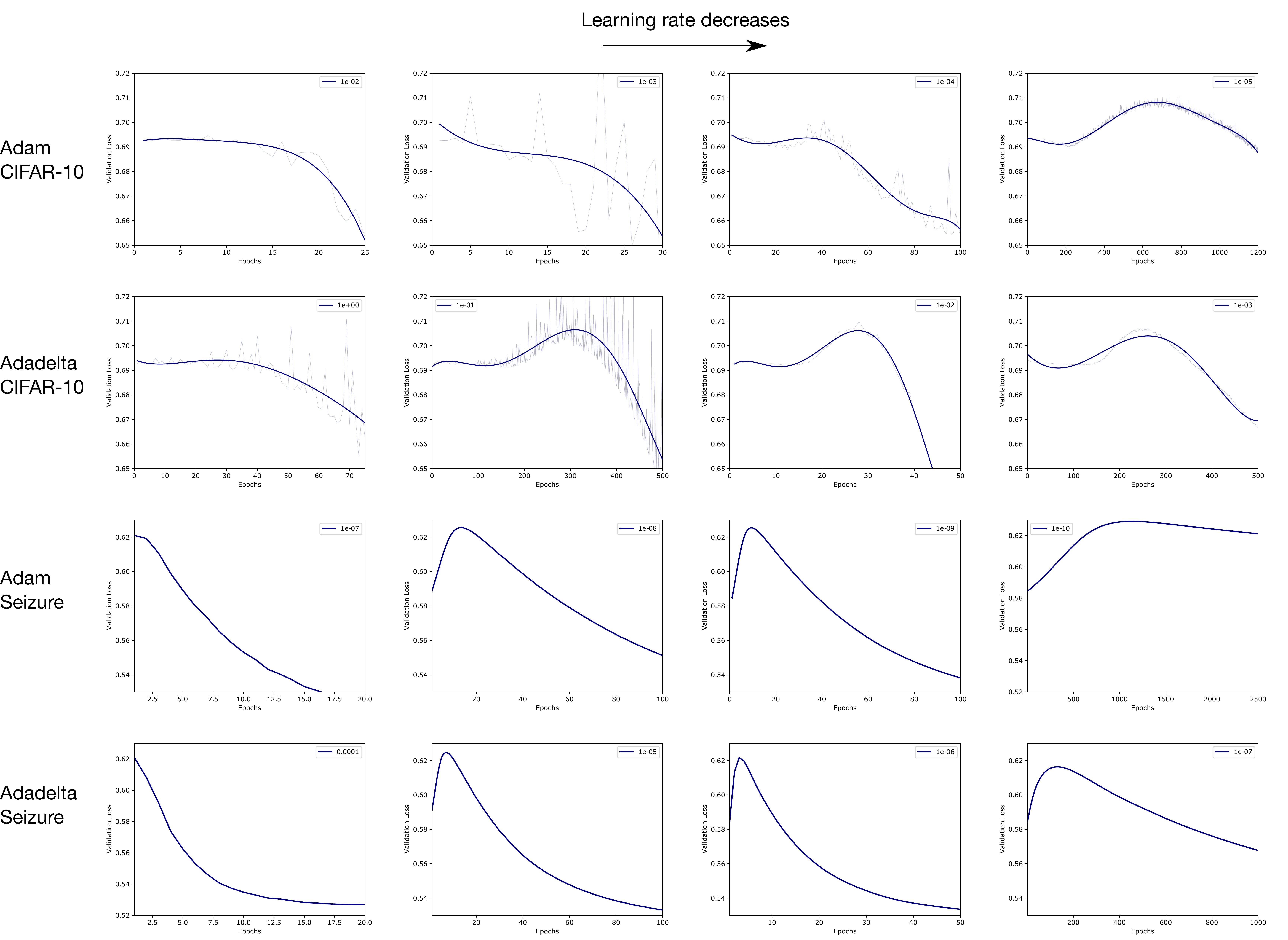}
  \caption{Influence of the learning rate on the apparition of a double descent pattern. Results are shown for both Adam and Adadelta optimizers. Both optimizers are susceptible to double descent. The two top rows are experiments on CIFAR-10, while the bottom rows correspond to experiments on seizure forecasting. For this experiment, a single patient was selected at random among patients for which a double descent pattern was observed. The learning rate decreases from right to left by increments of powers of 10, with its value indicated in the legends of each subplot. A polynomial was fitted to the estimated generalization error curve for the experiments on CIFAR-10, and not on the seizure forecasting. The generalization error is indicated on the y-axis as \textit{validation loss}. The boundaries of the x-axis differ per subplot. Plots with more epochs are given in supplementary materials.}
  \label{fig:lr_agg}
\end{figure}

\subsection{Influence of the learning rate}

We perform a series of experiments to assess the role of the optimization algorithm--the optimizer--in the apparition of double descent patterns. One of the main differences in modern optimizers, such as Adam \cite{kingma2014} or Adadelta \cite{zeiler2012}, is how the learning rate is computed. For example, Adadelta uses the exponentially decaying average of all past squared gradients as denominator and a unit normalization term as numerator. Adam uses the exponentially decaying average of all past gradients \textit{and} squared gradients as denominator and does not normalize the units. In addition to this automatic online tuning, a starting value for the learning rate needs to be set. The optimal value for this parameter usually depends on the task. Adam is widely used in deep learning and is known to be sensitive to the initial learning rate. In most libraries, Adam's default initial learning rate is set to 1e-3. On the contrary, Adadelta was designed to be insensitive to the initial learning, and the authors recommend keeping the default value $1$. We experiment both with Adam and Adadelta for varying learning rates.

At their default learning rate, both optimizers converge without double descent patterns on the CIFAR-10 task, but diverge on the seizure forecasting task (see Appendix). For the networks to converge, learning rates need to be decreased. Decreasing the order of magnitude of the learning rates makes double descent patterns appear progressively for both optimizers and both tasks (Figure \ref{fig:lr_agg}). In between high learning rates, which lead to direct divergence, and low learning rates, which lead to double descent patterns, there exist learning rates for which the networks converge and no double descent pattern is visible. The values of these learning rates are different depending on the optimizer and task (Figure \ref{fig:lr_agg}).

\subsubsection{Sampling rate and aliasing}

\begin{figure}[h!]
  \centering
  \includegraphics[width=14cm]{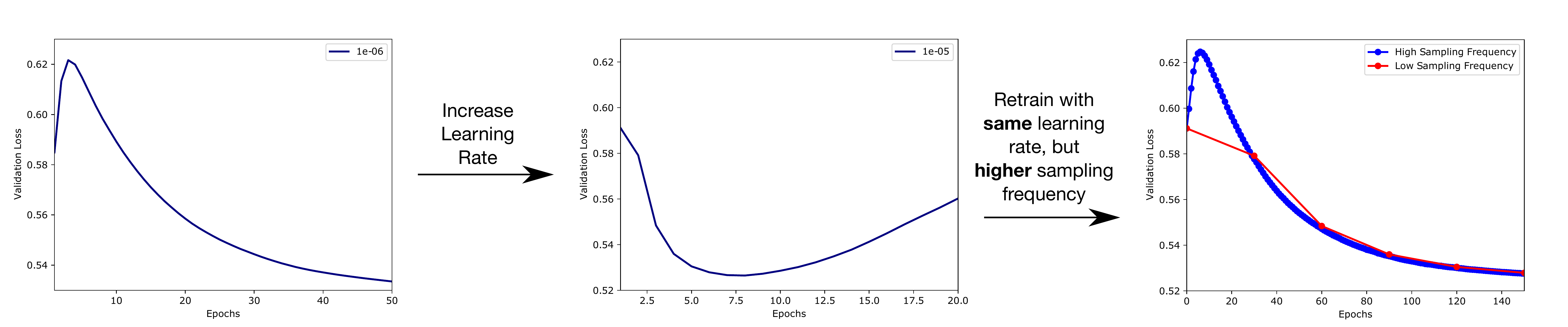}
  \caption{Aliasing can hide double descent patterns. Left: training with a learning rate of 1e-6. Middle: training with a learning rate of 1e-5. Right: increase of the sampling frequency when training with a learning rate of 1e-5. Both curves correspond to a different training. For the red curve, the generalization error is computed every 30 epoch instead of every epoch for the blue curve. All experiments are performed with Adadelta. The generalization error is indicated on the y-axis as \textit{validation loss}.}
  \label{fig:aliasing}
\end{figure}

We observed in the experiment above that increasing the learning can make the double descent pattern disappear. The double descent pattern may disappear because of two different reasons. The higher learning rate may either directly impact the true generalization error or only impact the estimation of generalization error and the subsequent polynomial fitting. Indeed, increasing the learning rate can partially be considered as decreasing the sampling of the estimated generalization error during training, which can create aliasing effects. The aliasing effect may mask the double descent pattern. Figure \ref{fig:aliasing} illustrates this phenomenon with real world data. The Figure shows two seizure forecasting experiments with the same optimizer (Adadelta) and learning rate corresponding to two networks training. For one of the trainings, the generalization error is estimated more often during training, which makes the double descent appear. Consequently, although we conclude above that higher learning rates can make the double descent disappear, the double descent pattern may still be present with higher learning rates but hidden by aliasing effects.

\section{Discussion}

We confirm earlier findings that double descent appears when the training set is both small and has noisy labels \cite{nakkiran2019}, and show that when one of these criteria is removed the generalization error converges monotonously in our experiments (Figures \ref{fig:trainsetsize} and \ref{fig:trainVSvalid}). In addition, we show that both training and generalization sets need to have noisy labels for the double descent pattern to appear.

We show that the order of magnitude of the learning rate has an impact on the apparition of a double pattern. Experiments on two datasets, with two different optimizer--Adam \cite{kingma2014} and Adadelta \cite{zeiler2012}--show that double descent appears with small learning rate and disappears when the order of magnitude of the learning rate is increased. We demonstrate that despite the recommendation of the authors, the learning rate of Adadelta \cite{zeiler2012} sometimes needs to be tuned to reach convergence of the generalization error, as exemplified by the experiments on seizure forecasting. Contrary to common understanding, when the training loss diverges, we do not recommend systematically decreasing the learning rate, as this may indicate the apparition of double descent pattern. Instead, one may want to also attempt increasing the learning rate.

We also show that increasing the learning rate may create an aliasing effect that hides the double pattern (Figure \ref{fig:aliasing}). We conclude that it is unclear whether increasing the order of magnitude of the learning effectively smooths the optimization landscape or simply hides the double descent pattern.

Nakkiran et al.\cite{nakkiran2019} hypothesized that the double descent pattern originates from an overparametrized model attempting to memorize the training data. While this setting is common in many areas of deep learning research, in which researchers attempt to use large models on their small local datasets, there are only a few reports of double descent patterns in the literature. We suspect that the double descent pattern occurs more frequently than reported, as it may be hidden by having both large learning rates and small sampling rates of the generalization error. 

Similarly, increasing the training set size may indirectly affect the frequency of the generalization error estimation during training and hide double descent patterns through aliasing effects. Indeed, epochs are defined as one pass over the full training dataset. If the training data is increased, and the generalization error is still estimated after each epoch, the sampling rate of the estimation of the generalization error is, in effect, reduced.

In some of the seizure experiments, the generalization error is smaller before the double descent pattern than after (Figure \ref{fig:multisub_sz}). We recommend ignoring the generalization error value pre-double descent, as it most probably corresponds to a model which memorized the training set and has consequently poor generalization capability. The model may by chance perform badly on the independent set gathered to compute the generalization error, and we believe that this would rather indicate that the generalization set is too small to estimate the true generalization error correctly.

In our experiments, we consider that there is no double descent pattern after the generalization error reaches a similar value as seen during the convergence of models trained with clean labels for CIFAR-10, or, for seizure forecasting, after the generalization error reaches a similar value as seen in other experiments, after the double descent pattern occurs. Yet, we continue training the networks much longer than necessary after convergence to verify that no double descent pattern appears later on. Following our own postulate, we still cannot be certain that any observed divergence of the generalization error does not indicate the start of a double descent pattern.

We did not address the influence of regularization techniques on the apparition of double descent patterns, although they can be expected to smooth the optimization landscape and, consequently, potentially remove double descent patterns. For example, Mei et al. \cite{mei2019} used data augmentation, $l_{1}$ regularization and $l_{2}$ regularization to smoothen the optimization landscape. Batch normalization \cite{ioffe2015} and Dropout \cite{srivastava2014} have also been shown to have a regularization effect and may also affect the apparition of double descent patterns.

\section{Conclusion}

We studied factors that influence the apparition of double descent patterns in the training of neural networks. We confirmed earlier findings that double descent appears when the training set is both small and has noisy labels, and showed that when one of these criteria was removed the generalization error converges monotonously in our experiments. In addition, we showed that both training and generalization sets needed to have noisy labels for the double descent pattern to appear. We showed that double descent appeared with small learning rates and disappeared when the order of magnitude of the learning rate was increased. We also showed that increasing the learning rate may have created an aliasing effect that hid the double descent pattern. We concluded that it is unclear whether increasing the order of magnitude of the learning effectively smooths the optimization landscape or simply hides the double descent pattern, and suspected that double descent patterns occurs more frequently than reported in the literature.

{
\small
\bibliographystyle{plain}
\bibliography{egbib}
}

\end{document}